\documentclass[review]{elsarticle}

\usepackage{lineno,hyperref}
\modulolinenumbers[5]
\usepackage{graphicx}
\usepackage{tikz}
\usepackage{comment}
\usepackage{amsmath,amssymb} 
\usepackage{color}
\usepackage{subcaption}
\usepackage{bbding}
\usepackage{multirow}
\usepackage{url}
\graphicspath{{figure/}}
\newcommand{\cmark}{\checkmark}%
\newcommand{\xmark}{\sffamily x}%

\journal{Pattern Recognition}

\begin{document}

\begin{frontmatter}

\title{How to Reduce Change Detection to Semantic Segmentation}

\author[mymainaddress]{Guo-Hua Wang}
\address[mymainaddress]{State Key Laboratory for Novel Software Technology, Nanjing University, China}

\author[mysecondaryaddress]{Bin-Bin Gao\corref{mycorrespondingauthor}}
\cortext[mycorrespondingauthor]{Corresponding author}
\ead{csgaobb@gmail.com}

\author[mysecondaryaddress,mythirdaddress]{Chenjie Wang}

\address[mysecondaryaddress]{Youtu Lab, Tencent, China}
\address[mythirdaddress]{Department of Computer Science and Engineering, Shanghai Jiao Tong University, China}

\begin{abstract}
Change detection (CD) aims to identify changes that occur in an image pair taken different times. Prior methods devise specific networks from scratch to predict change masks in pixel-level, and struggle with general segmentation problems. In this paper, we propose a new paradigm that reduces CD to semantic segmentation which means tailoring an existing and powerful semantic segmentation network to solve CD. This new paradigm conveniently enjoys the mainstream semantic segmentation techniques to deal with general segmentation problems in CD. Hence we can concentrate on studying how to detect changes. We propose a novel and importance insight that different change types exist in CD and they should be learned separately. Based on it, we devise a module named MTF to extract the change information and fuse temporal features. MTF enjoys high interpretability and reveals the essential characteristic of CD. And most segmentation networks can be adapted to solve the CD problems with our MTF module. Finally, we propose C-3PO, a network to detect changes at pixel-level. C-3PO achieves state-of-the-art performance without bells and whistles. It is simple but effective and can be considered as a new baseline in this field. Our code is at \url{https://github.com/DoctorKey/C-3PO}.
\end{abstract}


\end{frontmatter}

\section{Introduction}
\label{sec:intro}

\begin{figure}
	\centering
	\includegraphics[width=0.9\linewidth]{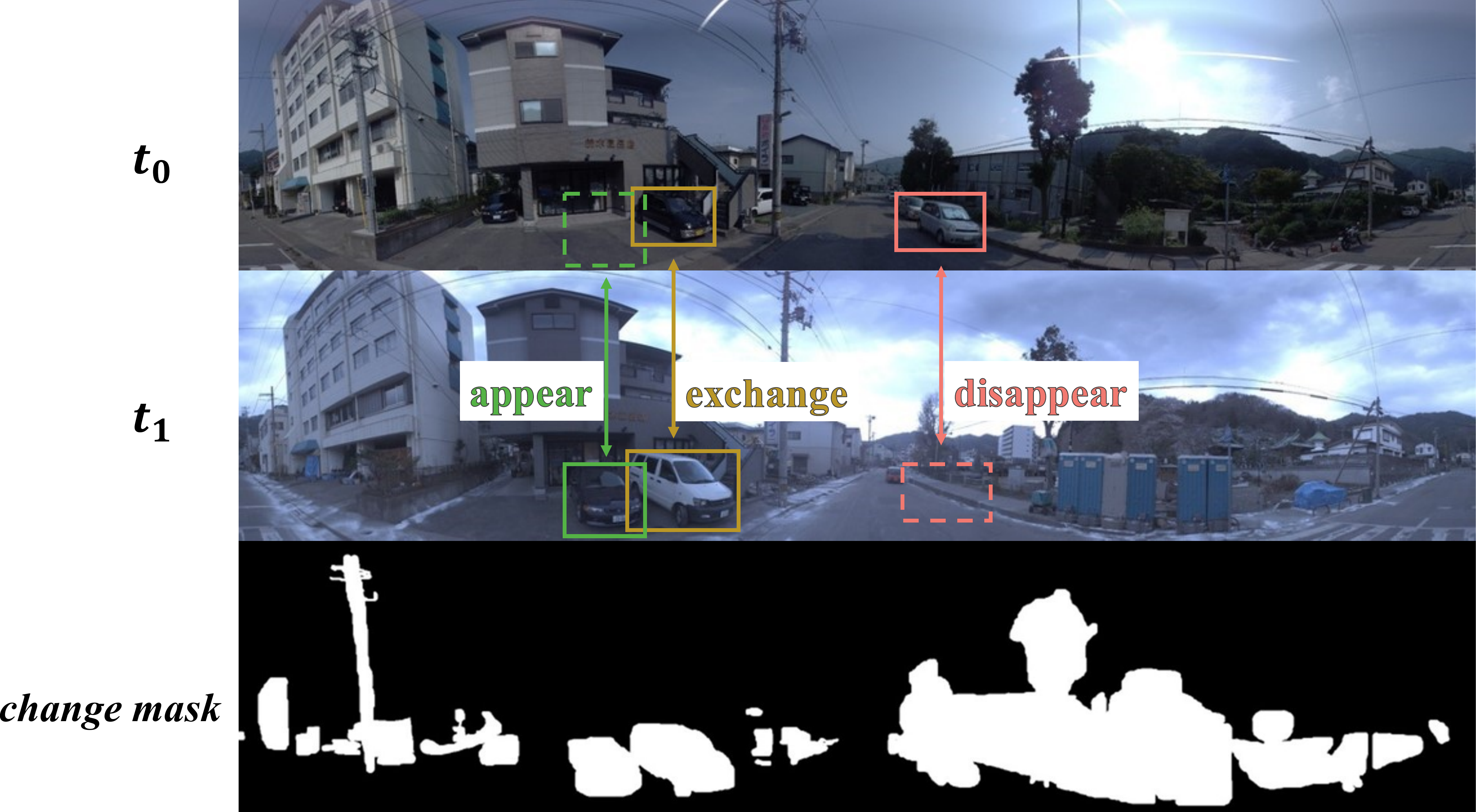}
	\caption{Illustration of change detection task. We propose there are three possible change types (i.e., ``appear'', ``disappear'' and ``exchange'', respectively) in this task. The network should model these three different change types separately.}
	\label{fig:CD}
\end{figure}

Change detection (CD) aims to identify changes that occur in a pair of images captured at different times as illustrated in Fig.~\ref{fig:CD}. A CD algorithm is required to predict change masks in pixel-level. It can be used in updating urban map, visual surveillance, disaster evaluation, agricultural industry and so on. Due to these various applications, CD has attracted more and more attention in the fields of computer vision.

\begin{figure}
	\centering
	\includegraphics[width=0.8\linewidth]{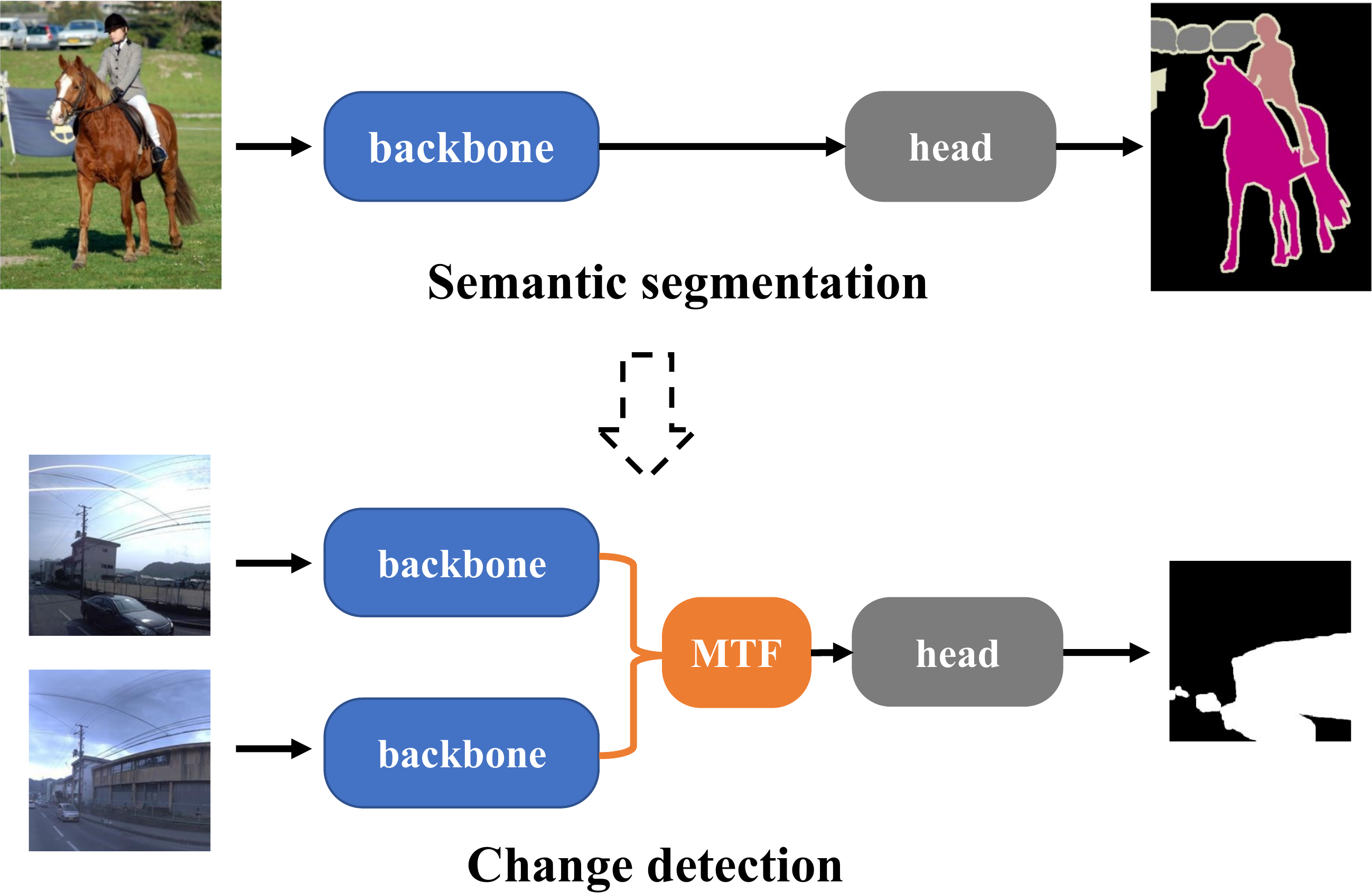}
	\caption{Illustration of our paradigm that reducing change detection to semantic segmentation. We propose the MTF module to fuse temporal features. The semantic segmentation network is transformed to solve CD by inserting MTF.}
	\label{fig:Trans}
\end{figure}

Recently, the deep neural network has achieved remarkable success in various computer vision tasks~\cite{wang2022versatile}. It is also adopted to deal with the CD task and achieves state-of-the-art performance. However, most researchers design specific networks from scratch to predict change masks and struggle with the general segmentation problems, such as how to deal with multi-scaled objects~\cite{HPCFNet} and how to improve the mask's details~\cite{DR_TANet}. In this paper, we decouple the change detection and the segmentation, and propose a new paradigm to solve CD as illustrated in Fig.~\ref{fig:Trans}. Thanks to it, a general semantic segmentation network can solve the CD problem effectively and efficiently. And general segmentation problems are alleviated by directly applying advantageous semantic segmentation networks. Compared with previous methods, the proposed paradigm make us get free of those segmentation problems so that we can concatenate on the core question: how to detect changes?

We propose a novel and essential insight that there are three possible change types in the CD task, and they should be learned separately. As illustrated by Fig.~\ref{fig:CD}, we name these changes as ``appear'', ``disappear'' and ``exchange'', respectively. These different change types are neglected by previous works (as the change mask in Fig.~\ref{fig:CD}). In this paper, we argue that the fusion feature should extract information of all these changes to help the segmentation head. We propose a module named MTF to \underline{M}erge two \underline{T}emporal \underline{F}eatures. And MTF is devised with a multi-branch structure and different branches focus on different change types. Experimental results demonstrate distinguishing three change types matters the performance significantly (cf. Tab.~\ref{tab:diff-branch}) and MTF indeed extracts these three changes (cf. Fig.~\ref{fig:branch}). Compared with previous fusion manners, our MTF enjoys high interpretability and reveals the essential characteristic of CD. And it is more generalized that can be easily inserted between the backbone and the segmentation head (as in Fig.~\ref{fig:Trans}).

At last, we propose C-3PO, namely \underline{C}ombine \underline{3} \underline{PO}ssible change types, to solve the change detection task. C-3PO utilizes the mature techniques of semantic segmentation. So it is reliable and easily deployed. With our MTF module, C-3PO can effectively solve the CD task with high interpretability. It outperforms previous methods and achieves state-of-the-art performance.

Our contributions can be summarized as follows.

$\bullet$ We propose a new paradigm to solve the CD problem. That is reducing change detection to semantic segmentation. Compared with previous paradigm that devises specific networks from scratch, our paradigm is more advantageous that it can make the most techniques in the fields of semantic segmentation.

$\bullet$ We argue that three possible change types are contained in the CD task, and they should be learned separately. The MTF module is proposed to combine two temporal features and extract all three change types. Our MTF is generalized well and can work with lots of segmentation networks. Experiments demonstrate the importance of distinguishing three change types and MTF indeed extracts information to identify them. 

$\bullet$ With our paradigm and MTF, we propose C-3PO, a network to detect changes in pixel-level. C-3PO is simple but effective. It achieves state-of-the-art performance without bells and whistles. It can be considered as a new baseline network in the CD field.

\section{Related Works}

\textbf{Change detection} requires the algorithm to identify changes according to image pairs. There are two scenarios in this field, i.e., remote-sensing change detection (RSCD) \cite{RSCD} and street scene change detection (SSCD)~\cite{PCD,VL_CMU_CD,changesim}. RSCD is based on remote sensing images and aims at information analysis on the earth's surface. \cite{ZHENG2022108717} proposes a high frequency attention Siamese network for a finer recognition of changed building objects in very-high-resolution remote sensed images. \cite{SUN2022108845} proposes an unsupervised iterative structure transformation and conditional random field based multimodal change detection method. SDACD~\cite{LIU2022108960} unifies image adaptation and feature adaptation in an end-to-end trainable manner, and handles cross-domain change detection. SSCD cares about street scene images. In this paper, we concentrate on SSCD because it has more applications. Early researches utilize handcrafted feature and feature matching to solve the CD problem~\cite{CD_survey,DOF}. Recently, the deep neural network is adopted and achieves better performance. However, most researchers design specific networks from scratch and struggle with the general segmentation problems. HPCFNet~\cite{HPCFNet} hierarchically combines features at multiple levels to deal with multi-scaled objects. Further, it introduces the MPFL~\cite{HPCFNet} strategy to alleviate the problem that the locations and the scales of changed regions are unbalanced. CosimNet~\cite{CosimNet} proposes a threshold contrastive loss to deal with the noise in the change masks. DR-TANet~\cite{DR_TANet} introduces the attention mechanism into the CD field and proposes CHVA~\cite{DR_TANet} to refine the strip entity changes. The performance of their methods is determined by both segmentation and change detection. And it is hard to distinguish which one contributes more. Although previous methods also use the framework of semantic segmentation for change detection. They consider these two problems at the same time, i.e., coupling semantic segmentation and CD. The previous paradigm can be considered as designing specific networks from scratch. In this paper, we propose a new paradigm that reducing CD to semantic segmentation, which means directly using a segmentation network to solve CD with minimal changes. Ours explicitly decouples CD into feature fusion and semantic segmentation. MTF is proposed and used to reduce CD to semantic segmentation, whereas previous methods cannot make this reduction explicitly. We can directly adopt a powerful semantic segmentation network, and therefore enjoy their techniques to deal with general segmentation problems. On other hand, we can concentrate on how to improve change detection, i.e., how to design and improve our MTF.

Previous works also study how to fuse two temporal features. The correlation layer is utilized by CSCDNet~\cite{CSCDNet} to estimate optical flow for localizing changes. HPCFNet \cite{HPCFNet} concatenates features and applies parallel atrous group convolution to capture multi-scaled information. Inspired by the self-attention mechanism, DR-TANet~\cite{DR_TANet} introduces temporal attention for the feature fusion. However, these fusion manners are heavily coupled with their specific network structure and hard to be applied with the general semantic segmentation networks. In this paper, we focus on how to build features for the general segmentation network to identify changes. We propose a novel and essential insight that distinguishing three change types will result in better performance. Finally, the proposed MTF module is more generalized that can be easily inserted in most semantic segmentation networks.

\textbf{Reduction} is one of the most important concepts in theoretical computer science, e.g. NP-Complete problems can be reduced to each other~\cite{Karp_Reducibility}. Transfer learning~\cite{transfer} can be considered as reduction in the fields of machine learning. When it comes to computer vision, RCNN~\cite{RCNN} reduces object detection to classification by extracting region proposals and classifying which proposal contains objects. FCOS~\cite{fcos} reduces object detection to semantic segmentation by predicting bounding boxes in a per-pixel fashion. ViT~\cite{ViT} reduces image classification to natural language processing (NLP) by splitting an image into patches and treating them as words in NLP. In this paper, we study how to reduce change detection to semantic segmentation. It is reasonable that both two tasks need to predict labels at the pixel level. And the techniques in the semantic segmentation field can indeed help to solve CD.

\textbf{Semantic segmentation} is a basic computer vision task that predicts the semantic classes on pixel-level. FCN~\cite{FCN} is an early work based on deep learning and achieve remarkable success in this task. Then a serial of DeepLab~\cite{ASPP,deeplabv3} are proposed and gradually become the mainstream approach. CENet~\cite{zhou2022contextual} introduces a novel encoder-decoder architecture to capture multi-scale context via ensemble deconvolution. GPNet~\cite{zhang2021gpnet} proposes a gated pyramid module to incorporate both low-level and high-level features.. Recently, attention-based models~\cite{DANet,ccnet} are introduced in this field, e.g. SETR~\cite{SETR} uses the transformer to deal with this task. In this paper, for robustness and reliability, we adopt FCN and DeepLab to solve the CD task. Note that with our proposed MTF module, it is easy to adjust recent segmentation networks to solve the CD problem.

\section{The Proposed Method}

\begin{figure}
	\centering
	\includegraphics[width=\linewidth]{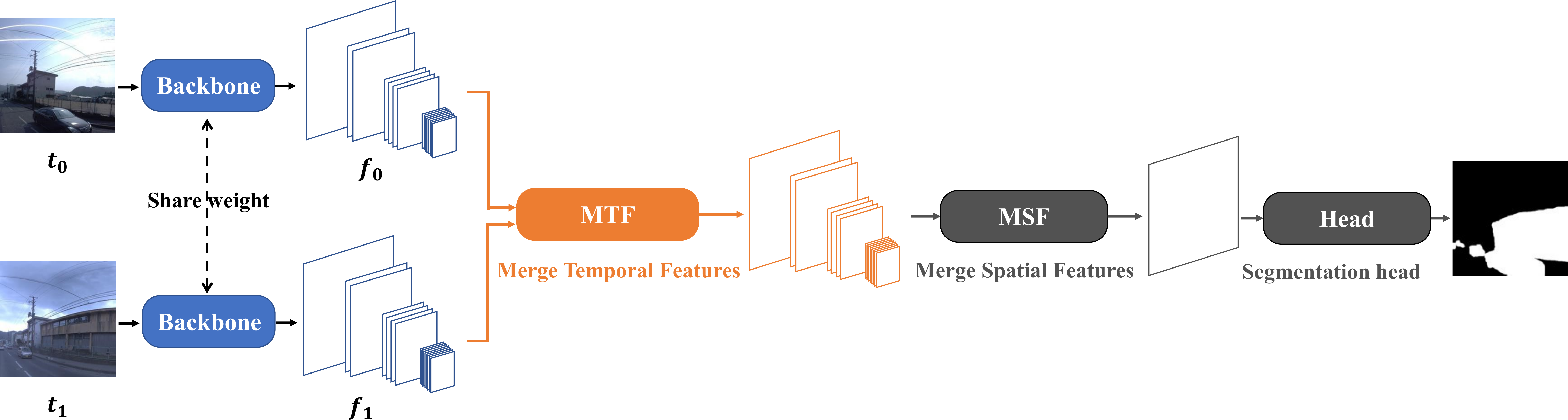}
	\caption{The overall architecture of C-3PO. Two temporal feature pyramids are extracted by the same backbone and merged by MTF. Then, MSF generates a single feature map for the segmentation head to predict change masks.}
	\label{fig:framework}
\end{figure}

Fig.~\ref{fig:framework} illustrates the overall architecture of our network. The main difference between our network and previous ones is that ours decouples change detection into feature fusion and semantic segmentation. Without the MTF module, the remained parts can form into a semantic segmentation network. So we can directly adopt the mainstream architectures of semantic segmentation in these parts. Our MTF module aims at extracting the requisite information from two temporal features for the general segmentation network to identify changes. The designation of MTF is based on our insight that different change types should be learned separately.

\textbf{Backbone:} A general-purpose backbone network is used to extract hierarchical feature maps in our framework. We adopt VGG~\cite{VGG}, ResNet~\cite{ResNet}, and MobileNetV2~\cite{mobilenet} in this paper. All backbones can generate pyramid features of multiple scales from 1/32 to 1/4 resolution. Given paired images $t_0$ and $t_1$, two types of pyramids $f_0$ and $f_1$ are extracted by the \emph{same} backbone, respectively. 

\begin{figure}
	\centering
	\includegraphics[width=0.8\linewidth]{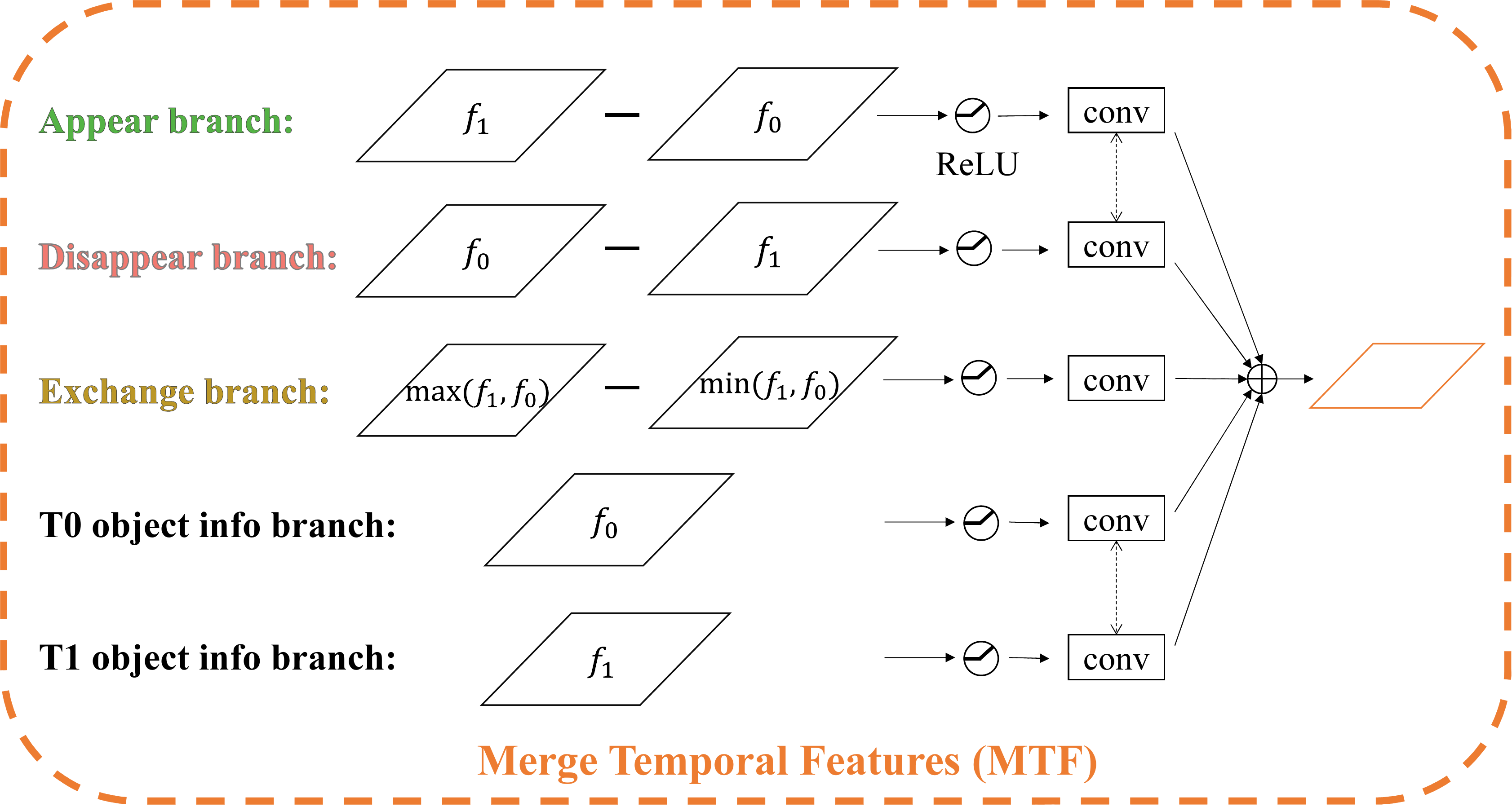}
	\caption{The detailed architecture of the MTF module. This module aims at merging two temporal feature maps into one. We use upper three branches to model ``appear'', ``disappear'' and ``exchange'', respectively.}
	\label{fig:MTF}
\end{figure}

\textbf{Merge temporal features (MTF):} Fig.~\ref{fig:MTF} shows the detailed architecture of our MTF module which has five branches. The upper three branches use the minus operation to model three different change types, while the under two branches directly use raw features to provide the basic semantic information of objects. ``Appear'' means the object only exists in $t_1$, so we expect $f_1$ is larger than $f_0$ in these positions. And we use $\text{ReLU}(f_1 - f_0)$ to model this change type. ``Disappear'' means the object only exists in $t_0$. Similar to ``Appear'', $\text{ReLU}(f_0 - f_1)$ is used in this branch. After the minus and ReLU operations, a conv. layer with kernel size $3\times 3$ is applied on these two branches. Because it is no need to distinguish the semantic changes that may exists in a specific image~(i.e., $t_0$ or $t_1$) and both two activation maps after ReLU identify there are objects only in one image, the ``Appear'' and ``Disappear'' branches share one conv. layer rather than customizing two different conv. layers.

``Exchange'' means objects in $t_0$ and $t_1$ are different. This change can be considered as a object in $t_0$ disappears and a new object appears in $t_1$. To this end, we define the exchange as
\begin{align}
  \label{eq:exchange}
  \text{Exchange} &= \text{Disappear} + \text{Appear} \\
  &= \text{ReLU}(f_0 - f_1) + \text{ReLU}(f_1 - f_0) \\
  &= \max(f_0, f_1) - \min(f_0, f_1)\,.
\end{align}
This formulation is reasonable due to the change mask of exchange type is according to more significant objects between $t_0$ and $t_1$ (cf. Fig.~\ref{fig:CD}). Finally, a $3 \times 3$ conv. layer is added after the activation map. 

The under two branches directly use $f_0$ and $f_1$ to provide the object's semantic information. Objects (e.g. a car) can be detected by the extracting pyramid raw features. The upper three branches will lose this semantic information due to the minus operations. We find that it needs the under two branches to identify objects for detecting changes better.

At last, all branches' outputs are summarized into a single feature map. We force the shape of the output feature map is the same as that of $f_0$ and $f_1$ by properly setting the conv. layers' channels. So our MTF can be easily inserted between the backbone and the segmentation head.

With the multi-branch structure, our MTF enjoys sufficient expressivity. Concatenating and subtracting two features are two widely used fusion manners in the fields of change detection~\cite{FC_EF,ChangeNet,HPCFNet,CosimNet}. The under two branches in MTF can model the concatenating fusion while the exchange branch computes the absolute difference between two features. So our MTF is more expressive than most previous fusion manners.

\begin{figure}
	\centering
	\includegraphics[width=0.8\linewidth]{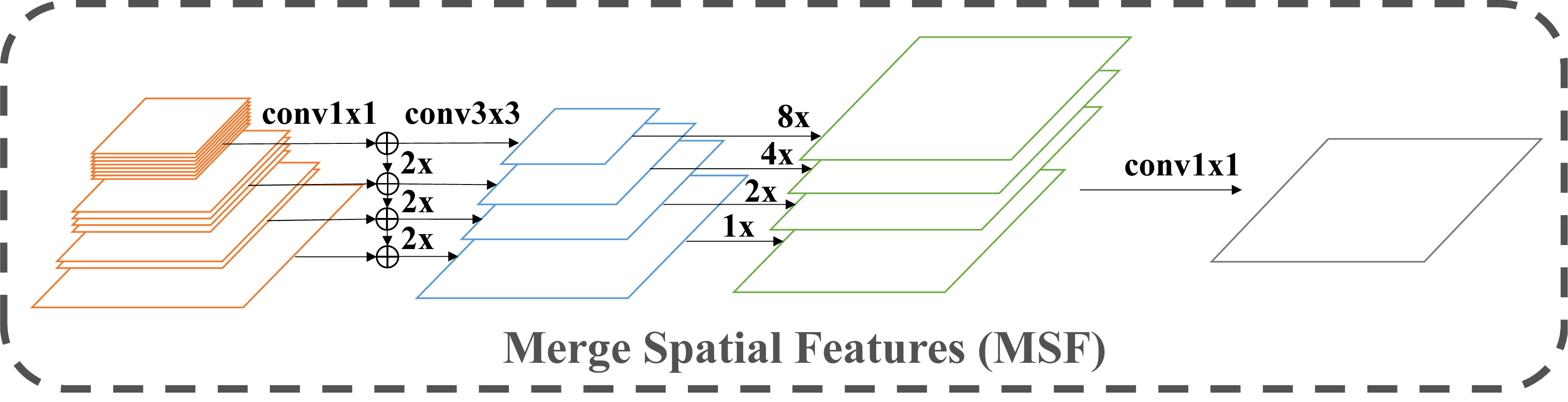}
	\caption{The detailed architecture of the MSF module. This module aims at merging the pyramid into a single feature map.}
	\label{fig:MSF}
\end{figure}

\textbf{Merge spatial features (MSF):} Fusing different spatial features can effectively deal with the multi-scaled change objects~\cite{HPCFNet}. Different from previous works that designing complex architectures by themselves, we directly use FPN~\cite{FPN} and make minimal changes to merge spatial features. The FPN architecture has been verified by lots of works~\cite{PFPN,mask_rcnn}. Fig.~\ref{fig:MSF} shows its detailed architecture. Given a pyramid with 4 spatial scales (i.e., 1/4, 1/8, 1/16 and 1/32), all feature maps are reduced to $256$ channels by $1\times 1$ conv. firstly. Then the smaller feature maps are enlarged by $2\times$ bilinear upsampling and added in the lower feature map. $3\times 3$ conv. are used to further process these feature maps. To construct the final single feature map, the upper 3 feature maps are enlarged by bilinear upsampling into the resolution of the lowest feature map. All feature maps are concatenated and reduced to $512$ channels by $1\times 1$ conv.

\textbf{Semantic segmentation head:} Once constructing the final feature map, we can adopt the existing semantic segmentation heads to predict the change mask. In this paper, FCN~\cite{FCN} and DeepLabv3~\cite{deeplabv3} are used. The FCN head first uses $3\times 3$ conv. to reduce the number of channel by 4 times, and then uses $1\times 1$ conv. to further reduce the channel to $N$ classes. Finally, $4\times$ bilinear upsampling and softmax are used to generate the prediction mask at the original image resolution. Compared with FCN, DeepLabv3 is more advantageous that consists of ASPP~\cite{ASPP} and can effectively capture multi-scale information. 

\subsection{Training}

The change detection task requires the model to predict a change mask to indicate the change region. This can be considered as a semantic segmentation task with multi classes. Hence, the softmax cross-entropy loss function is used to train our network. However, the distribution of change/background is heavily imbalanced (e.g. $0.06:0.94$ in VL-CMU-CD task). Following previous works~\cite{HPCFNet}, we use the weighted cross-entropy loss to alleviate this problem. Assume there are total $N$ classes from $0$ to $N - 1$ (the $0$-th class denotes the background), and the loss function on each pixel is defined as
\begin{equation}
  \label{eq:loss}
  \mathcal{L} = - \sum_{i=0}^{N-1} w_i y_i \log p_i\,,
\end{equation}
where $p_i$ and $y_i$ are the prediction and ground-truth of the $i$-th class region, respectively. $w_i = \frac{\sum_{j=0}^{N-1} n_j - n_i}{(N-1) \sum_{j=0}^{N-1} n_j}$ is the balance weight, where $n_i$ is the number of $i$-th class pixels. We use the training set to calculate these statistics. Finally, the loss averaged over pixels is used for optimization. 

Typically, only two classes (i.e., change/background) are considered in the CD field. Recently, a dataset with multi-class is proposed~\cite{changesim}. Note that our method can solve binary-class and multi-class CD problems with Eq.~\ref{eq:loss}.

\subsection{Analysis}
\label{sec:analysis}

\textbf{``Appear'' v.s. ``Disappear'':} The concepts of ``Appear'' and ``Disappear'' are relative. As illustrated in Fig.~\ref{fig:CD}, it can be considered as a car appears in the green box. On the other hand, it can also be considered as the ground disappears. Although the latter opinion seems weird (due to humans often focus on objects), these two opinions are both make sense for the change detection task. When it comes to deep neural networks, C-3PO can surprisingly deal with these two opinions. That means given a dataset which only contains ``disappear'' change, C-3PO can work well with the disappear branch or appear branch. More experimental details can be found in Section~\ref{sec:merge_t}.

\textbf{Symmetry:} Symmetry is another important property of our framework. Obviously, the features generated by our MTF module will not change if we change the order of image pairs (i.e., exchange $t_0$ and $t_1$). This symmetry results from the sharing weights backbones and the symmetrical architecture of the MTF module. In MTF, we use the same conv. layer after the appear/disappear branches, and the same conv. layer after two info branches. Hence, exchanging image pairs' order will keep the output unchange. The symmetry implicitly augments training data by exchanging their orders. It is reasonable because exchanging the order should not affect the algorithm to identify changes in an image pair. So this symmetry property can alleviate the overfitting problem and boost performance. More experimental results can be found in Section~\ref{sec:symmetry}. 

\textbf{Structure transfer:} Note that our MTF module is versatile to apply in most semantic segmentation frameworks. That means given a semantic segmentation network, it can be used for solving CD tasks by inserting our MTF module. So we in fact propose a framework to generate CD networks. And the performance can be boosted by applying more powerful segmentation networks.

\textbf{Parameter transfer:} Without the MTF module, the remained parts can form into a semantic segmentation network. This network can be firstly trained on semantic segmentation tasks. Then, the parameters can be transferred to the CD task. Note that it is difficult for previous CD networks to decompose into segmentation networks and be pretrained on the semantic segmentation task.

\section{Experimental Results}

In this section, first, we introduce datasets, implementation details, and evaluation metrics. Then, a series of ablation studies are conducted to study our method. Finally, we compare C-3PO with state-of-the-art methods.

\textbf{VL-CMU-CD} is proposed in \cite{VL_CMU_CD}. This task \emph{only} contains ``disappear'' semantic changes. Following previous works, we resize images into $512\times 512$ resolution and split the dataset into training and testing. The training set consists of $933$ image pairs, while the testing set consists of $429$ pairs. Following previous works~\cite{CSCDNet,DR_TANet,FC_EF,ChangeNet}, the training set is augmented to $3732$ pairs by rotation. We train models on the training set and report evaluation metrics on the testing set.

\textbf{PCD} is proposed in \cite{PCD} and three change types (i.e., ``appear'', ``disappear'' and ``exchange'') exist in it (cf. Fig.~\ref{fig:CD}). PCD can be divided into two subsets (i.e., ``GSV'' and ``TSUNAMI''), and each subset has $100$ image pairs and hand-labeled change mask. The resolution of each image is $224\times 1024$. Following previous works~\cite{DR_TANet,HPCFNet}, we crop the original images by sliding $56$ pixels in width. Hence, $3000$ paired patches with $224\times 224$ resolution are generated. Then, these patches are augmented by rotate $0^{\circ}$, $90^{\circ}$, $180^{\circ}$ and $270^{\circ}$. In total, $12000$ paired patches are generated. Finally, 5-fold cross-validation is adopted for training and testing. Specifically, each training set consists of $9600$ paired patches with $224\times 224$ resolution, while each testing set consists of $20$ raw paired images with $224\times 1024$ resolution. Following previous works, we reshape images to $256\times 1024$ for testing. 

\textbf{ChangeSim} is proposed in \cite{changesim} and all three change types exist in it. The training and testing set consist of $13225$ and $8212$ pairs, respectively. Following \cite{changesim}, the images are firstly resized to $256\times 256$, and then horizontal flipping and color jittering are used for the data augmentation. We simply resize images to $256\times 256$ for testing.

\textbf{Implementation details.} To train the network, we use Adam~\cite{Adam} with an initial learning rate of $0.0001$. And the learning rate is decayed by the cosine strategy. We use $4$ V100 GPUs with batch size $16$ (i.e., $4$ images per GPU). For a fair comparison, the backbone is pretrained on ImageNet~\cite{imagenet} while other parts are initialized randomly except experiments in Section~\ref{sec:transfer}. All networks are trained for $100$ epochs. Note that \emph{all} experiments use the same training strategy and there are \emph{no} more hyperparameters to tune.

\begin{figure}
	\centering
	\includegraphics[width=0.7\linewidth]{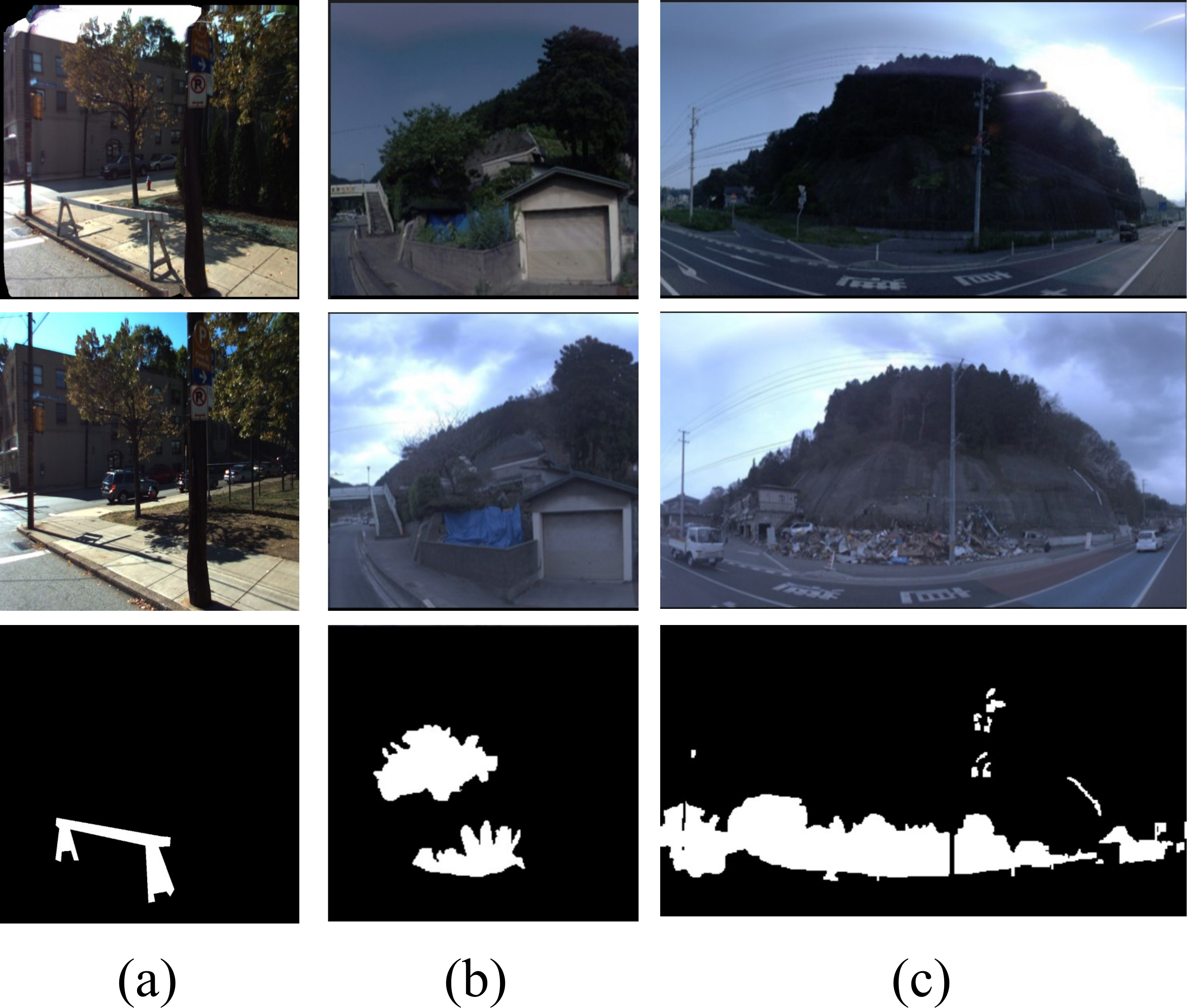}
	\caption{Examples of noisy label. Images in (a) are from VL-CMU-CD and the changed cars are not annotated. Examples in (b) and (c) are from PCD and the changed trees are not annotated well.}
	\label{fig:noise}
\end{figure}

\textbf{Evaluation metrics.} Following previous works~\cite{DR_TANet,HPCFNet}, we evaluate the performance on VL-CMU-CD and PCD with F1-score. It is calculated upon Precision and Recall. In these two binary CD task, the change regions represent positive while the background (unchanged regions) represent negative. Given true positive (TP), false positive (FP) and false negative (FN), the F1-score is defined as
\begin{equation}
  \text{F1-score}=\frac{2\times \text{Precision} \times \text{Recall}}{\text{Precision} + \text{Recall}}\,,
\end{equation}
where $\text{Precision}=\frac{TP}{TP+FP}$ and $\text{Recall}=\frac{TP}{TP+FN}$. In addition, some ground-truth masks are not accurate in these change detection tasks (cf. Fig.~\ref{fig:noise}). Following the researches on noise label~\cite{PENCIL,dividemix}, we evaluate the network after each epoch and report both best and last performance. 

ChangeSim contains two CD tasks (i.e., binary and multi-class). Following~\cite{changesim}, we evaluate the performance with mIoU, which is defined as 
\begin{equation}
  \text{mIoU}=\frac{1}{C}\sum^C_{i=1}\frac{P_i\cap G_i}{P_i\cup G_i}\,,
\end{equation}
where $C$ is the total number of category, and $P_i$ and $G_i$ denote the prediction and groundtruth masks for the $i$-th class.

\subsection{Merge temporal features}
\label{sec:merge_t}

\begin{figure}
	\subcaptionbox{\label{fig:sub1}}{\includegraphics[width=0.48\linewidth]{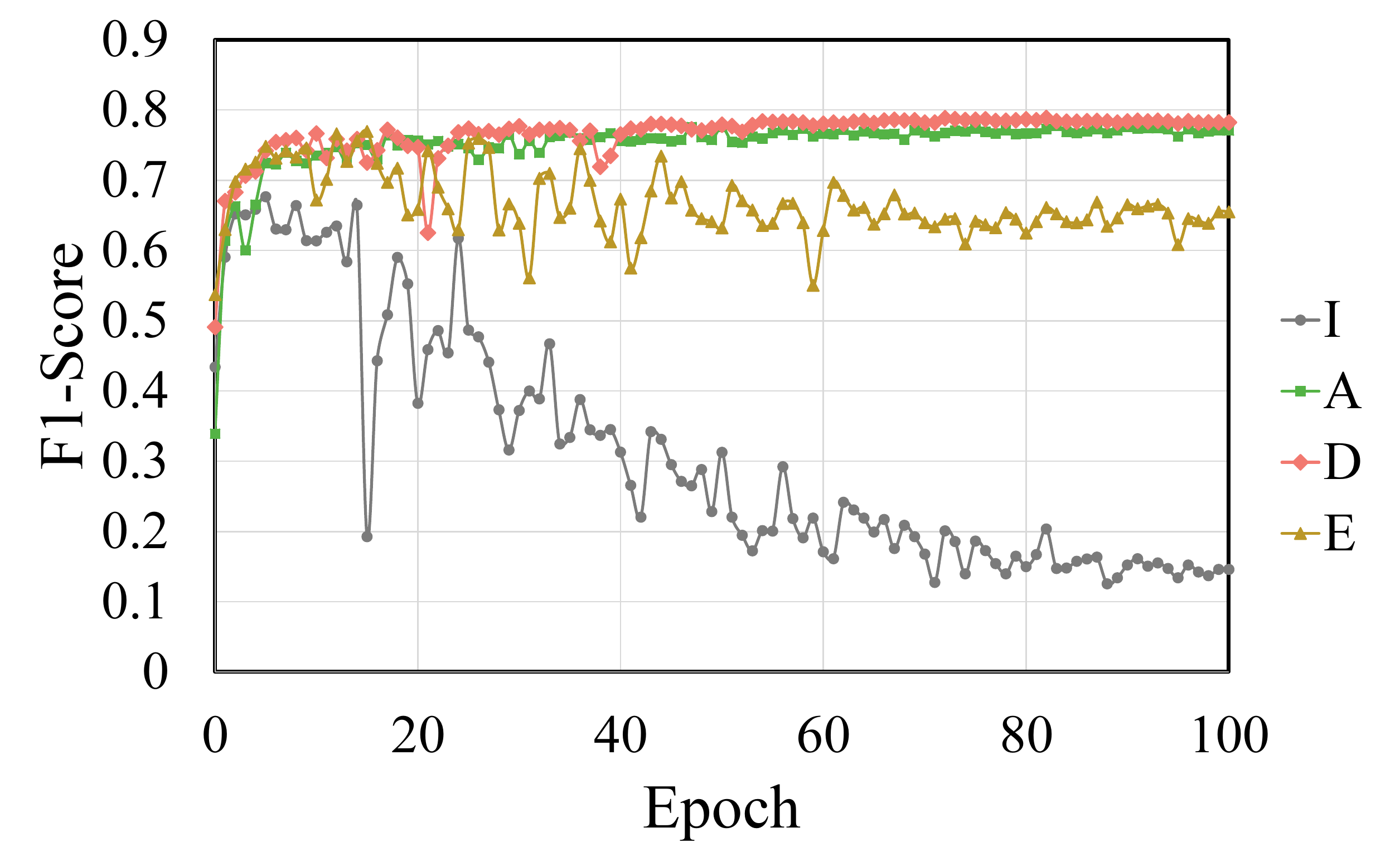}}
	\subcaptionbox{\label{fig:sub2}}{\includegraphics[width=0.48\linewidth]{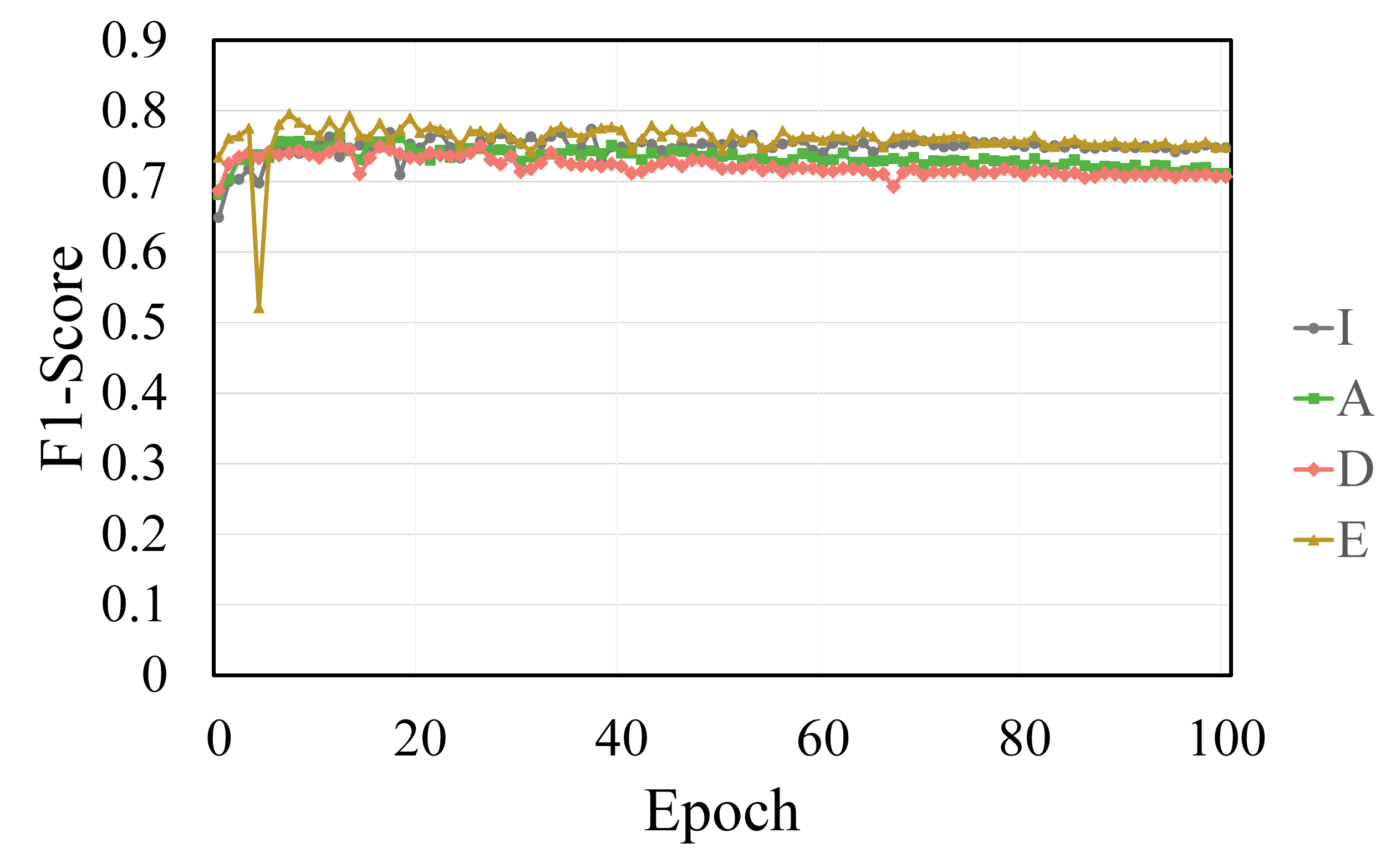}}
\caption{F1-score of C-3PO after each training epoch. (\protect\subref{fig:sub1}) presents the results on VL-CMU-CD which only contains the ``disappear'' change. (\protect\subref{fig:sub2}) shows the results on GSV which contains three change types. ``I'', ``A'', ``D'' and ``E'' denote the info, appear, disappear and exchange branches in Fig.~\ref{fig:MTF}, respectively.}
\label{fig:CMU_plot} 
\end{figure}

To evaluate our insight that different change types should be learned separately, we conduct ablative experiments with using different branches in the MTF module.Fig.~\ref{fig:sub1} shows F1-score on VL-CMU-CD for C-3PO with just one type branch after each epoch. Because this task only contains ``disappear'' changes, both appear and disappear branches perform well according to analysis in Section~\ref{sec:analysis}. The info branch loses the discrepant information to detect changes. And the exchange branch struggles with the dilemma that detecting changes according to objects or background (i.e., whether to make the object or background feature larger during training). These results demonstrate the manner of feature fusion is needed to suit the task.

\begin{figure}
	\centering
	\includegraphics[width=\linewidth]{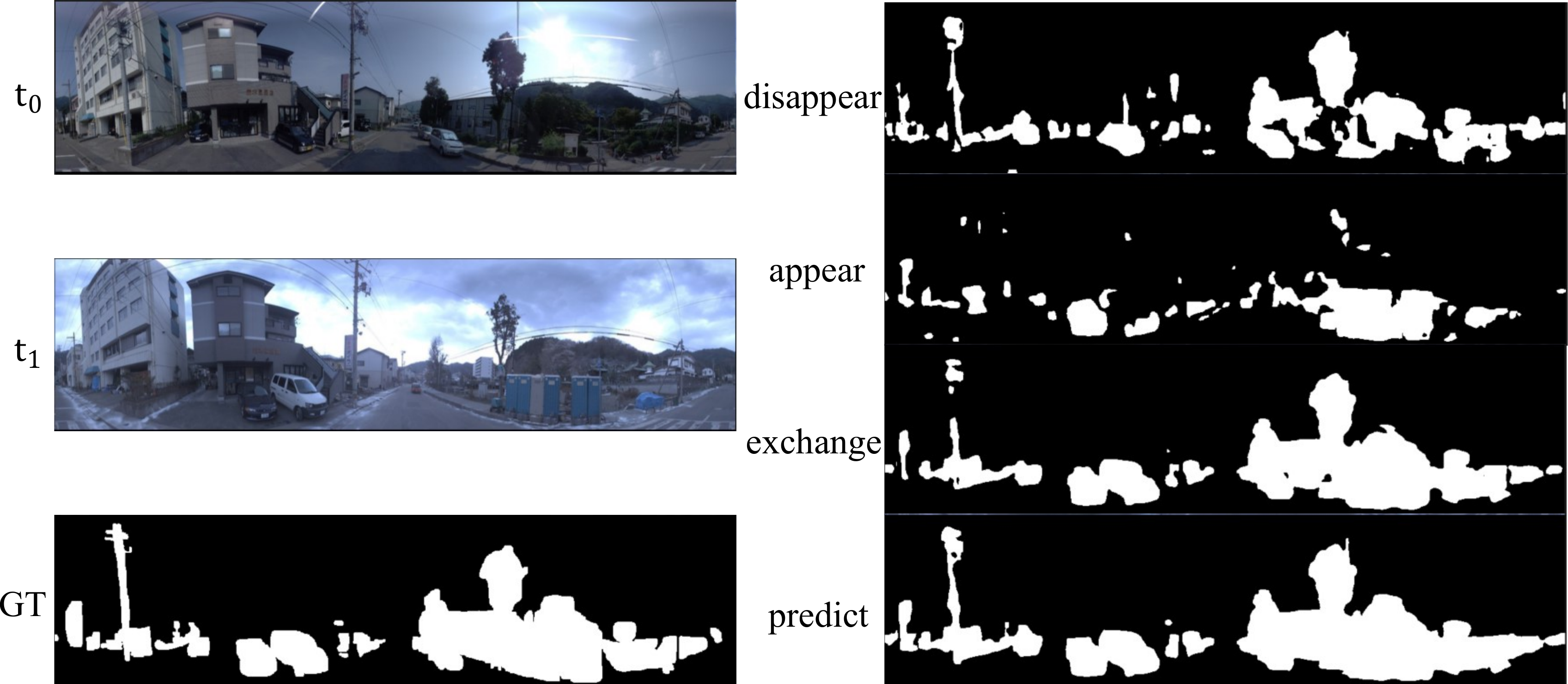}
	\caption{Visual illustration of change masks predicted by C-3PO with different branches. We train C-3PO with all three change branches, then use just one branch to identify changes. ``disappear'', ``appear'' and ``exchange'' denote three change branches, respectively. ``predict'' uses all branches and ``GT'' is the ground-truth change mask.}
	\label{fig:branch}
\end{figure}

GSV is different from VL-CMU-CD that it contains all three change types. Fig.~\ref{fig:sub2} shows results. First, the exchange branch performs best among all three change branches, because three change types exist in this task. Second, because changes in GSV are heavily based on objects, C-3PO works well even with only the info branches. For example, as in Fig.~\ref{fig:CD}, cars are changed with a high probability in image pairs. Third, it is unsuitable to only use the unidirectional change branch such as appear and disappear.

Fig.~\ref{fig:branch} shows the change masks predicted by C-3PO with different branches. Three change branches indeed extract different features to identify different change types. The prediction of ``exchange'' is close to the final prediction due to it combines appear and disappear (cf. Equation~\ref{eq:exchange}). So our MTF enjoys high interpretability and reveals the essential characteristic of CD.

\setlength{\tabcolsep}{5pt}
\begin{table}
  \centering
  \small
  \caption{F1-score (\%) on benchmarks for C-3PO with using different branches in the MTF module. ``I'', ``A'', ``D'' and ``E'' denote the info, appear, disappear and exchange branches in Fig.~\ref{fig:MTF}, respectively.}
  \begin{tabular}{ccccccc}
    \hline
    \multirow{2}*{Structure} & \multicolumn{2}{c}{VL-CMU-CD} & \multicolumn{2}{c}{GSV} & \multicolumn{2}{c}{TSUNAMI}  \\
     & last & best & last & best & last & best \\
    \hline
    I & $14.6$ & $67.7$ & $75.2$ & $76.9$ & $83.7$ & $84.6$ \\
    A & $77.1$ & $77.7$ & $73.0$ & $76.3$ & $84.1$ & $85.0$  \\
    D & $\mathbf{78.2}$ & $\textbf{78.8}$ & $73.4$ & $76.2$ & $84.2$ & $85.1$  \\
    E & $65.5$ & $77.0$ & $\textbf{76.4}$ & $\textbf{79.1}$ & $\textbf{84.7}$ & $\textbf{85.7}$  \\
    \hline
    I + A & $\textbf{78.8}$ & $\textbf{79.6}$ & $75.4$ & $77.6$ & $84.6$ & $85.4$  \\
    I + D & $78.3$ & $79.4$ & $75.9$ & $77.9$ & $84.9$ & $\textbf{85.9}$  \\
    I + E & $59.4$ & $74.6$ & $\textbf{76.6}$ & $\textbf{79.4}$ & $\textbf{85.0}$ & $\textbf{85.9}$  \\
    \hline
    I + A + E & $77.7$ & $\textbf{79.3}$ & $76.7$ & $78.9$ & $84.9$ & $85.9$  \\
    I + D + E & $\textbf{78.2}$ & $79.2$ & $76.8$ & $78.9$ & $84.9$ & $86.0$  \\
    I + A + D & $57.0$ & $77.2$ & $\textbf{77.0}$ & $\textbf{79.4}$ & $\textbf{85.1}$ & $\textbf{86.1}$ \\
    \hline
    I + A + D + E & $53.7$ & $76.5$ & $\textbf{77.6}$ & $\textbf{79.4}$ & $84.7$ & $85.4$ \\
    \hline
  \end{tabular}
  \label{tab:diff-branch}
\end{table}

\begin{table*}
  \centering
  \small
  \caption{F1-score (\%) for C-3PO with different positions to merge temporal information and different pretrained models. $\odot_{T}$ and $\odot_{S}$ denotes temporal and spatial fusions, respectively. $B$ and $H$ denotes backbone and semantic segmentation head, respectively. ImageNet pretraining means B is pretrained on ImageNet, while COCO pretraining means $B$, $\odot_{S}$ and $H$ are pretrained on COCO.}
  \begin{tabular}{cccccccc}
    \hline
    \multirow{2}*{Structure} & \multirow{2}*{Pretrain} & \multicolumn{2}{c}{VL-CMU-CD} & \multicolumn{2}{c}{GSV} & \multicolumn{2}{c}{TSUNAMI} \\
     & & last & best & last & best & last & best \\
    \hline
    \multirow{2}*{$\odot_{T} \rightarrow B \rightarrow \odot_{S} \rightarrow H$} & ImageNet & $69.1$ & $70.6$ & $43.7$ & $53.2$ & $63.1$ & $67.8$ \\
    & COCO & $70.2$ & $72.1$ & $47.4$ & $52.5$ & $63.6$ & $67.2$  \\
    \hline
    \multirow{2}*{$B \rightarrow \odot_{T} \rightarrow \odot_{S} \rightarrow H$} & 
    ImageNet & $77.6$ & $79.4$ & $77.6$ & $\textbf{79.4}$ & $84.7$ & $85.4$ \\
    & COCO & $\textbf{78.9}$ & $\textbf{79.9}$ & $\textbf{77.8}$ & $78.8$ & $\textbf{84.8}$ & $85.5$ \\
    \hline
    \multirow{2}*{$B \rightarrow \odot_{S} \rightarrow \odot_{T} \rightarrow H$} & ImageNet & $76.4$ & $76.7$ & $75.3$ & $77.7$ & $84.7$ & $\textbf{85.7}$ \\
    & COCO & $78.3$ & $78.8$ & $75.9$ & $77.7$ & $84.6$ & $85.4$ \\
    \hline
    \multirow{2}*{$B \rightarrow \odot_{S} \rightarrow H \rightarrow \odot_{T}$} & ImageNet & $7.4$ & $12.6$ & $55.9$ & $61.9$ & $73.3$ & $78.0$ \\
    & COCO & $67.3$ & $72.6$ & $60.2$ & $64.1$ & $74.7$ & $77.7$ \\
    \hline
  \end{tabular}
  \label{tab:position_MTF}
\end{table*}

Tab.~\ref{tab:diff-branch} presents the results of C-3PO with different branch combinations. On VL-CMU-CD, comparing ``I + A/D'' and ``A/D'', we find adding object information is beneficial to detect change. ``I + E'' still works badly due to adding object information cannot solve the dilemma. And ``I + A + D'' achieves a similar performance to ``I + E'' because we define the exchange by Equation~\ref{eq:exchange}. Finally, ``I+A/D'' performs best among all combinations. Note that VL-CMU-CD only contains the disappeared changes. Models should have the right inductive bias for the unidirectional changes. As discussed in Sec.~\ref{sec:analysis}, both the appear and disappear branches have the inductive bias that makes C-3PO detect changes according to objects or backgrounds. When it comes to GSV and TSUNAMI, these two datasets contain all change types. Hence, the model should be equipped with all change branches, especially the exchange branch for the bidirectional changes. Considering the difference among these benchmarks, in the rest of this paper we use ``I + D'' on VL-CMU-CD, and ``I + A + D + E'' on GSV, TSUNAMI, and ChangeSim. Note that these structures are not the best combinations in Tab.~\ref{tab:diff-branch}. We choose the model's structure according to the training set rather than the performance on the testing set. From training sets, VL-CMU-CD only contains disappeared changes, while other datasets have all change types. Our choices achieve comparable performance with the best combinations (less than 1\% F1-score). We believe these choices are reasonable and give sound conclusions in the latter experiments.

\subsection{Temporal fusion position}

We study where to merge temporal information and Tab.~\ref{tab:position_MTF} summarizes the experimental results. We placed the MTF module in four different positions. ``$\odot_{T} \rightarrow B \rightarrow \odot_{S} \rightarrow H$'' means combining image pairs by MTF directly. Premature merging results in overfitting easily, because it does not utilize the backbone's ability to detect objects. Inserting MTF between the backbone and MSF achieves the best performance. And placing MTF after MSF also works well. These two settings utilize the backbone to detect objects and the head to predict change masks. Merging temporal information before spatial information can make the MSF module obtain more information for detecting changes. Finally, placing MTF after the head expects the head can predict the semantic mask of each image and the change mask can be predicted by simply combining two semantic masks. That is difficult due to only change masks are available to train the network.

\subsection{Parameter transfer}
\label{sec:transfer}

We evaluate the performance of C-3PO with COCO~\cite{COCO} pretraining and Tab.~\ref{tab:position_MTF} summarizes the results. As introduced in Section~\ref{sec:analysis}, $B$, $\odot_{S}$ and $H$ can form into a general semantic segmentation network which can be pretrained on COCO. On VL-CMU-CD, C-3PO with COCO pretraining outperforms that with ImageNet pretraining by a significant margin. We find the boost is largest with $B \rightarrow \odot_{S} \rightarrow H \rightarrow \odot_{T}$ structure. With this structure, $B \rightarrow \odot_{S} \rightarrow H$ is pretrained, and it has the ability to semantically segment. Hence, it can work well that adding $\odot_{T}$ to directly merge two semantic maps. On PCD, the training set is sufficient to train the segmentation head.  

\subsection{Symmetry}
\label{sec:symmetry}

\begin{table}
  \centering
  \small
  \caption{F1-score (\%) for C-3PO with sharing specific parts. ``B'', ``I'' and ``C'' denote the backbone, the info branches and the appear/disappear branches, respectively. ``\cmark'' denotes sharing weights. Note that we use ``I+D'' on VL-CMU-CD, so it does not need to consider the symmetry for appear/disappear branches in this model. We evaluate networks on the first cross-validation fold.}
  \begin{tabular}{ccccccccc}
    \hline
    \multicolumn{3}{c}{Share} & \multicolumn{2}{c}{VL-CMU-CD} & \multicolumn{2}{c}{GSV} & \multicolumn{2}{c}{TSUNAMI} \\
    B & I & C & last & best & last & best & last & best \\
    \hline
    \xmark & \xmark & \xmark & $79.5$ & $80.2$  & $72.2$ & $76.3$ & $88.2$ & $89.0$\\
    \xmark & \cmark & \cmark & $79.4$ & $80.2$  & $72.4$ & $76.5$ & $88.3$ & $88.9$\\
    \cmark & \xmark & \xmark & - & - & $73.0$ & $77.0$ & $88.7$ & $89.2$ \\
    \cmark & \cmark & \xmark & - & - & $74.8$ & $77.7$ & $88.5$ & $89.1$ \\
    \cmark & \xmark & \cmark & $\textbf{79.9}$ & $\textbf{80.6}$ & $74.7$ & $77.4$ & $88.2$ & $89.1$ \\
    \cmark & \cmark & \cmark & $79.5$ & $80.2$ & $\textbf{76.5}$ & $\textbf{79.3}$ & $\textbf{88.8}$ & $\textbf{89.4}$ \\
    \hline
  \end{tabular}
  \label{tab:symmetry}
\end{table}

The symmetry of C-3PO is due to the sharing weights used in the backbone, the info branches and the appear/disappear branches. We conduct experiments for C-3PO without sharing specific parts and Tab.~\ref{tab:symmetry} summarizes results. The symmetry matters significantly on GSV and sharing weights in all three parts achieves the best performance. Without the symmetry, it is easy for the network to overfit the individual information in two input images (e.g., the weather in each image). Using a symmetrical framework can make the network concentrate on the changes in the image pair. And it implicitly augments the image pair by exchanging their order which further alleviates the overfitting problem.

\subsection{Weighted loss}
\label{sec:weighted-loss}

\begin{table}
  \centering
  \small
  \caption{F1-score (\%) for C-3PO with/without weighted loss. ``\cmark'' denotes using balance weight in Eq.~\ref{eq:loss}. We evaluate networks on the first cross-validation fold.}
  \begin{tabular}{ccccccc}
    \hline
    \multirow{2}*{weighted loss} & \multicolumn{2}{c}{VL-CMU-CD} & \multicolumn{2}{c}{GSV} & \multicolumn{2}{c}{TSUNAMI} \\
     & last & best & last & best & last & best \\
    \hline
     & $77.4$ & $79.3$ & $\textbf{76.8}$ & $\textbf{79.7}$ & $88.3$ & $89.1$ \\
    \cmark & $\textbf{79.5}$ & $\textbf{80.2}$ & $76.5$ & $79.3$ & $\textbf{88.8}$ & $\textbf{89.4}$ \\
    \hline
  \end{tabular}
  \label{tab:weighted-loss}
\end{table}

Following previous works~\cite{DR_TANet,CSCDNet,HPCFNet}, we adopt the weighted cross-entropy loss to alleviate the imbalance problem. Tab.~\ref{tab:weighted-loss} makes a comparison for the loss without weighting. This balanced weight matters significantly on VL-CMU-CD, because its change/background distribution is about $0.06:0.94$, more imbalanced than that of PCD ($0.29:0.71$). Overall, as argued by previous works, we also appreciate using weighted cross-entropy loss on these tasks.

\subsection{More configurations}
\label{sec:conf}

\begin{table}
  \centering
  \small
  \caption{F1-score (\%) for C-3PO with different configurations of the backbone, MSF and the head.}
  \begin{tabular}{ccccccccc}
    \hline
    \multirow{2}*{Backbone} & \multirow{2}*{MSF} & \multirow{2}*{Head} & \multicolumn{2}{c}{VL-CMU-CD} & \multicolumn{2}{c}{GSV} & \multicolumn{2}{c}{TSUNAMI} \\
     & & & last & best & last & best & last & best \\
    \hline
    ResNet-18 & 1 & FCN & $73.4$ & $73.6$ & $71.1$ & $73.6$ & $82.1$ & $82.5$  \\
    ResNet-18 & 2 & FCN & $77.9$ & $78.3$ & $75.7$ & $77.8$ & $84.1$ & $84.9$ \\
    ResNet-18 & 3 & FCN & $78.1$ & $79.2$ & $76.9$ & $78.5$ & $\textbf{85.1}$ & $\textbf{85.7}$ \\
    ResNet-18 & 4 & FCN & $\textbf{79.5}$ & $\textbf{80.2}$ & $\textbf{77.2}$ & $\textbf{79.0}$ & $85.0$ & $\textbf{85.7}$ \\
    \hline
    ResNet-18 & 1 & DeepLabv3 & $72.7$ & $73.1$ & $71.7$ & $74.0$ & $81.7$ & $82.3$ \\
    ResNet-18 & 2 & DeepLabv3 & $\textbf{79.0}$ & $79.1$ & $76.6$ & $78.1$ & $84.3$ & $84.8$ \\
    ResNet-18 & 3 & DeepLabv3 & $\textbf{79.0}$ & $79.3$ & $77.3$ & $79.2$ & $\textbf{84.9}$ & $\textbf{85.6}$  \\
    ResNet-18 & 4 & DeepLabv3 & $77.6$ & $\textbf{79.4}$ & $\textbf{77.6}$ & $\textbf{79.4}$ & $84.7$ & $85.4$ \\
    \hline
    MobileNetV2 & 4 & DeepLabv3 & $77.8$ & $79.3$ & $\textbf{76.9}$ & $78.8$ & $84.9$ & $85.8$ \\
    ResNet-50 & 4 & DeepLabv3 & $77.6$ & $79.9$ & $76.5$ & $\textbf{80.0}$ & $85.1$ & $86.2$  \\
    VGG-16 & 4 & DeepLabv3 & $\textbf{79.6}$ & $\textbf{80.0}$ & $76.8$ & $79.5$ & $\textbf{85.6}$ & $\textbf{86.5}$ \\
    \hline
  \end{tabular}
  \label{tab:conf}
\end{table}

To demonstrate the generalization of our paradigm, we evaluate C-3PO with different configurations of the backbone, MSF and semantic segmentation head. Tab.~\ref{tab:conf} summarizes the results. First, the number of MSF denotes how many feature maps are used in the MSF module. ``1'' means only using the feature map with spatial scale 1/32. The F1-score gets higher with adding more low-level feature maps. The low-level feature maps contain more details for C-3PO to segment more accurately. Second, we try FCN and DeepLabv3 as the semantic segmentation head. Both FCN and DeepLabv3 perform well on these benchmarks. Third, different backbones are adopted. The backbone with a larger capacity results in better performance.

\subsection{Different viewpoints}
\label{sec:viewpoints}

\begin{figure}
	\centering
	\includegraphics[width=0.8\linewidth]{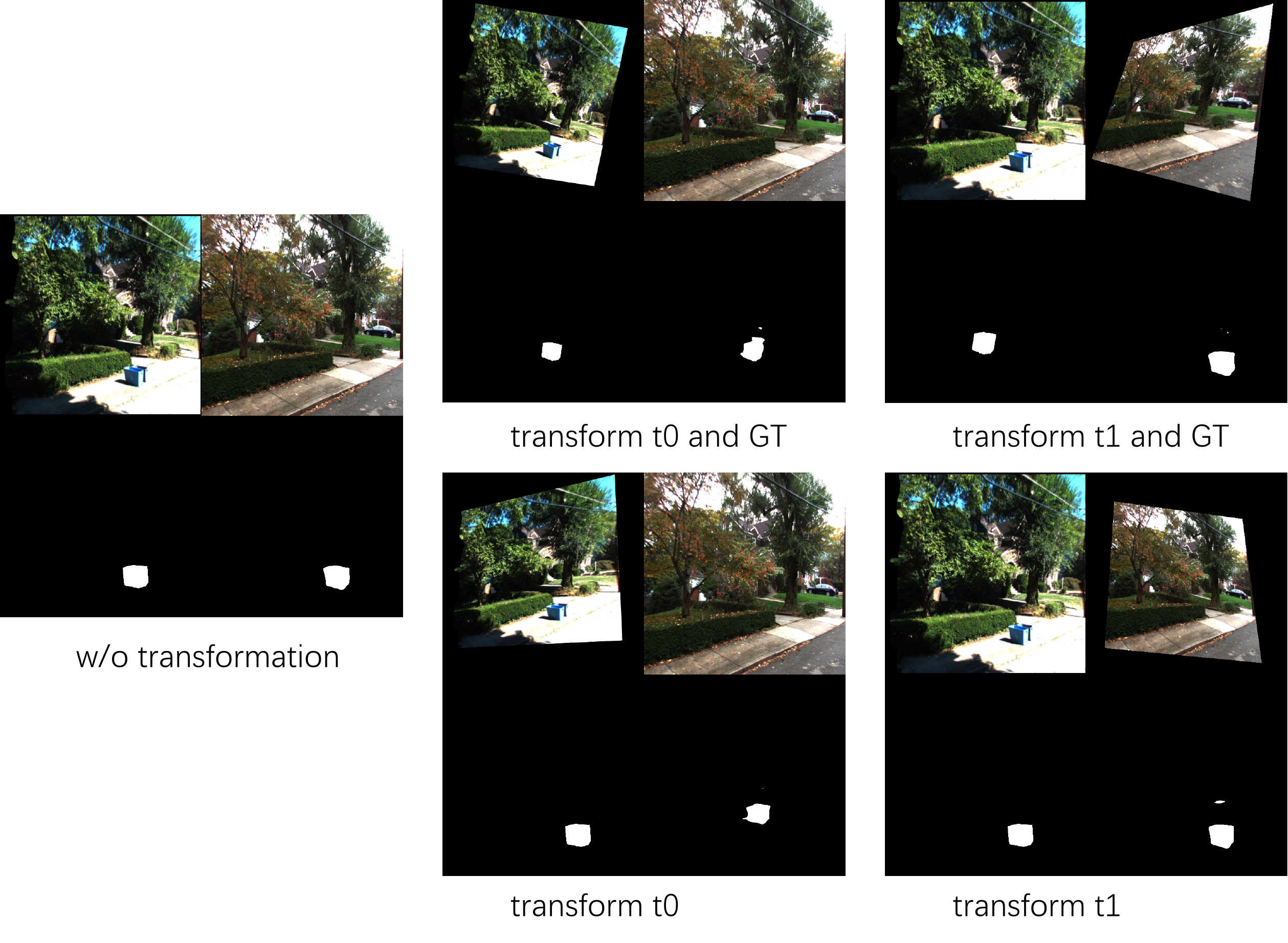}
	\caption{Visualization of C-3PO on testing images with perspective transformations.}
	\label{fig:perspective}
\end{figure}

In the case of street scene change detection, the viewpoints of two input images may differ slightly. We perform the perspective transformation of the input images to study this scenario. Fig.~\ref{fig:perspective} presents visualization results. In each sub-figure, the top two images are the images captured at $t_0$ and $t_1$, respectively. The bottom two masks are the ground truth and the model's prediction, respectively. We transform one image with keeping another unchanged. Note that all predictions are from the same model trained without this transformation augmentation. First, to evaluate the accuracy, the crucial problem is how to define the groundtruth (GT) change mask, i.e., choosing which viewpoint for it. We find that it easily results in misalignment of the model's prediction and the GT mask. Our model predicts according to the viewpoint of $t_0$, because the disappeared object exists in this image. And our model is robust that detecting this object in all cases. Overall, we suggest adopting techniques from multi-view geometry to align two images first to alleviate the perspective problem, and therefore avoid the dilemma of how to define GT masks.

\subsection{Different object sizes}
\label{sec:size}

\begin{table}
  \centering
  \small
  \caption{Results for C-3PO on VL-CMU-CD w.r.t. different object sizes.}
  \begin{tabular}{c|ccc|c}
    \hline
     & {Small} & {Medium} & {Large} & {All} \\
    \hline
    Ratio (\%) & $0\sim 2$ & $2\sim 6$ & $6\sim$ & $0\sim $\\
    \#Data & 133 & 157 & 139 & 429 \\
    \hline
    F1-score (\%) & 76.1 & 80.3 & 81.4 & 79.3 \\
    \hline
  \end{tabular}
  \label{tab:size}
\end{table}

\begin{figure}
	\centering
	\includegraphics[width=0.8\linewidth]{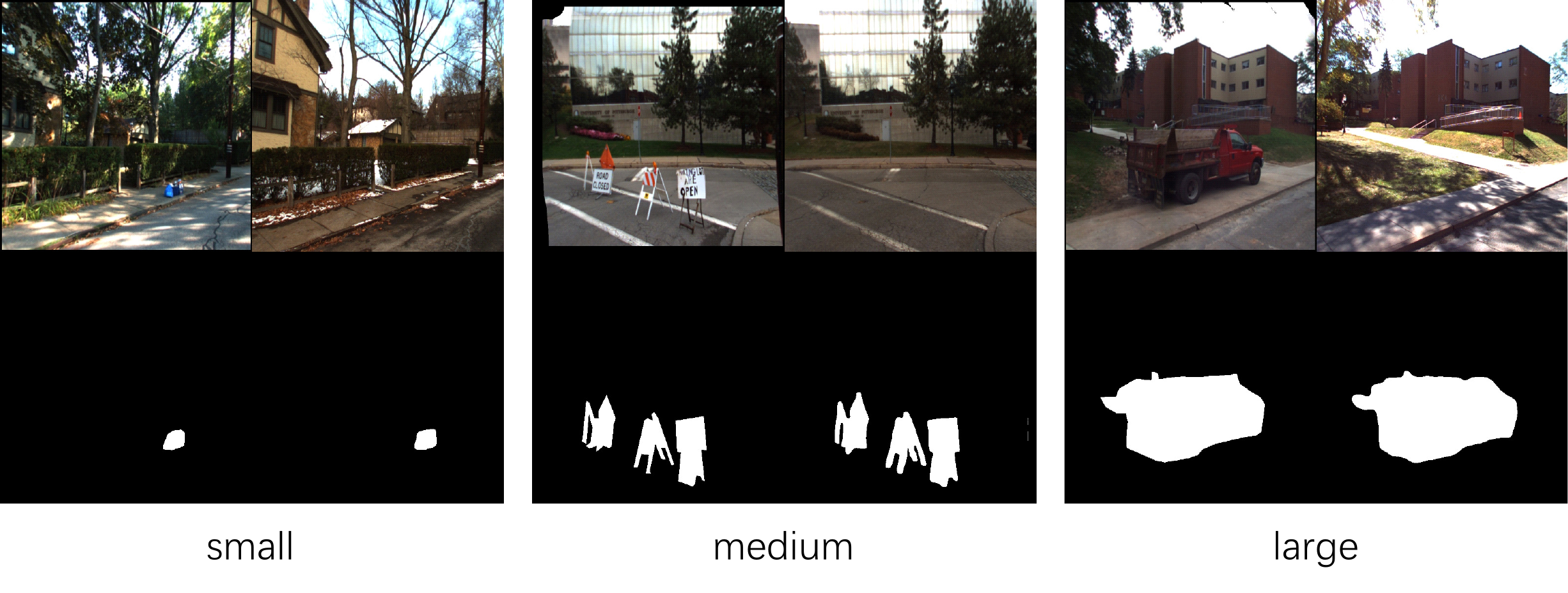}
	\caption{Visualization of C-3PO on testing images with different object sizes. }
	\label{fig:object_size}
\end{figure}

To study the performance w.r.t. the changing object size, we split the testing set of VL-CMU-CD into three folds. Given the binary change masks, we calculate the ratio for the changing area over the whole area. According to this ratio, ``Small'', ``Medium'', and ``Large'' testing sets are obtained (cf. Tab.~\ref{tab:size}) with comparable data amounts. F1-scores are evaluated on each testing set. As shown in Tab.~\ref{tab:size}, it is more challenging to detect small objects. Fig.~\ref{fig:object_size} visualizes results with different sizes. Overall, our model detects all objects successfully.

\subsection{Compared with state-of-the-art}
\label{sec:sota}

\begin{table}
  \centering
  \small
  \caption{F1-score (\%) for C-3PO and previous methods on VL-CMU-CD.}
  \begin{tabular}{lccc}
    \hline
    \multirow{2}*{Method} & \multirow{2}*{Backbone} & \multicolumn{2}{c}{VL-CMU-CD} \\
     & & last & best \\
    \hline
    FC-EF~\cite{FC_EF} & U-Net & 43.2 & 44.6 \\
    FC-Siam-diff~\cite{FC_EF} & U-Net & 65.1 & 65.3 \\
    FC-Siam-conc~\cite{FC_EF} & U-Net & 65.6 & 65.6 \\
    DR-TANet~\cite{DR_TANet} & ResNet-18 & 72.5 & 75.1 \\
    CSCDNet~\cite{CSCDNet} & ResNet-18 & 75.4 & 76.6 \\
    C-3PO & ResNet-18 & $77.6$ & $\textbf{79.4}$ \\
    \hline
    HPCFNet~\cite{HPCFNet} & VGG-16 & \multicolumn{2}{c}{75.2} \\
    C-3PO & VGG-16 & $79.6$ & $\textbf{80.0}$ \\
    \hline
  \end{tabular}
  \label{tab:sota_cmu}
\end{table}

\begin{table}
  \centering
  \small
  \caption{F1-score (\%) for C-3PO and previous methods on PCD.}
  \begin{tabular}{lccccccc}
    \hline
    \multirow{2}*{Method} & \multirow{2}*{Backbone} & \multicolumn{2}{c}{GSV} & \multicolumn{2}{c}{TSUNAMI} & \multicolumn{2}{c}{Average} \\
     & & last & best & last & best & last & best \\
    \hline
    FC-EF~\cite{FC_EF} & U-Net & $63.3$ & $64.7$ & $77.1$ & $77.7$ & $70.2$ & $71.2$ \\
    FC-Siam-diff~\cite{FC_EF} & U-Net & $64.0$ & $66.2$ & $78.6$ & $79.5$ & $71.3$ & $72.8$ \\
    FC-Siam-conc~\cite{FC_EF} & U-Net & $69.8$ & $70.4$ & $81.2$ & $81.6$ & $75.5$ & $76.0$ \\
    DR-TANet~\cite{DR_TANet} & ResNet-18 & $72.0$ & $74.3$ & $83.0$ & $84.5$ & $77.5$ & $79.4$ \\
    CSCDNet~\cite{CSCDNet} & ResNet-18 & $71.1$ & $75.0$ & $83.2$ & $84.8$ & $77.1$ & $79.9$ \\
    C-3PO & ResNet-18 & $77.6$ & $\textbf{79.4}$ & $84.7$ & $\textbf{85.4}$ & $81.2$ & $\textbf{82.4}$\\
    \hline
    HPCFNet~\cite{HPCFNet} & VGG-16 & \multicolumn{2}{c}{77.6} & \multicolumn{2}{c}{\textbf{86.8}} & \multicolumn{2}{c}{82.2}\\
    C-3PO & VGG-16 & $76.8$ & $\textbf{79.5}$ & $85.6$ & $86.5$ & $81.2$ & $\textbf{83.0}$ \\
    \hline
  \end{tabular}
  \label{tab:sota_pcd}
\end{table}

Finally, we compare C-3PO with state-of-the-art approaches. Tab.~\ref{tab:sota_cmu} and \ref{tab:sota_pcd} summarize the results. For a fair comparison, C-3PO \emph{only} utilizes the ImageNet pretrained backbones as previous works. DeepLabv3 is adopted to be head in C-3PO. We rerun the code of state-of-the-arts supplied by authors and report the F1-score of the last and best epoch. Note that HPCFNet~\cite{HPCFNet} does not release the code, and it is difficult to reproduce their method. So we directly cite their numbers. On VL-CMU-CD, C-3PO outperforms CSCDNet and HPCFNet by $2.8\%$ and $4.8\%$, respectively. That is a significant improvement compared with previous works. With the FCN head, the performance can be further boosted by $0.8\%$. On the PCD benchmark, C-3PO outperforms CSCDNet and HPCFNet by $2.5\%$ and $0.8\%$, respectively. 

Finally, we evaluate C-3PO on the ChangeSim benchmark. Tab.~\ref{tab:changesim} summarizes the results. The ChangeSim benchmark contains two tasks. The multi-class setting requires the model to classify different change types, while the binary setting expects the model to generate the binary change mask. In the multi-class setting, ``new'' and ``missing'' can be roughly considered as ``appear'' and ``disappear'' in our paper, respectively. And all other change types (``change'', ``rotated'', and ``replaced'') are considered as ``exchange'' in our model. We simply set different numbers of final output channels according to different settings. At last, C-3PO outperforms other methods on both two tasks by a significant margin. Although ours improves the state-of-the-art results on this benchmark, it is still challenging and needs more effort to overcome the difficulties. Different from PCD and VL-CMU-CD, ChangeSim contains more change types, i.e., ``rotated''. To imporve the performance on ChangeSim, a more sophisticated MTF module should be proposed. However, this paper focuses on the general model for all change detection datasets. We leave the sophisticated model for only ChangeSim as the future work.

\begin{table}
  \centering
  \small
  \caption{The mIoU (\%) performance on the ChangeSim benchmark for C-3PO and previous methods. ``S'', ``C'', ``N'', ``M'', ``Ro'' and ``Re'' denote ``static'', ``change'', ``new'', ``missing'', ``rotated'' and ``replaced'', respectively.}
  \begin{tabular}{cccccccccc}
    \hline
    \multirow{2}*{Method} & \multicolumn{3}{c}{Binary} & \multicolumn{6}{c}{Multi-class} \\
     & S & C & mIoU & S & N & M & Ro & Re & mIoU \\
    \hline
    ChangeNet~\cite{ChangeNet} & $73.3$ & $17.6$ & $45.4$ & $80.6$ & $9.1$ & $6.9$ & $11.6$ & $6.6$ & $23.0$ \\
    CSCDNet~\cite{CSCDNet} & $87.3$ & $22.9$ & $55.1$ & $90.2$ & $12.4$ & $6.0$ & $17.5$ & $7.9$ & $26.8$ \\
    C-3PO & $90.4$ & $28.8$ & $\textbf{59.6}$ & $92.6$ & $13.3$ & $8.0$ & $16.9$ & $8.0$ & $\textbf{27.8}$ \\
    \hline
  \end{tabular}
  \label{tab:changesim}
\end{table}

Overall, our C-3PO outperforms state-of-the-art by a significant margin. And due to its simplicity, we believe it can be served as a new baseline in this field and inspire more researches.

\subsection{Visualization}
\label{sec:visual}

\begin{figure}
	\centering
	\includegraphics[width=0.8\linewidth]{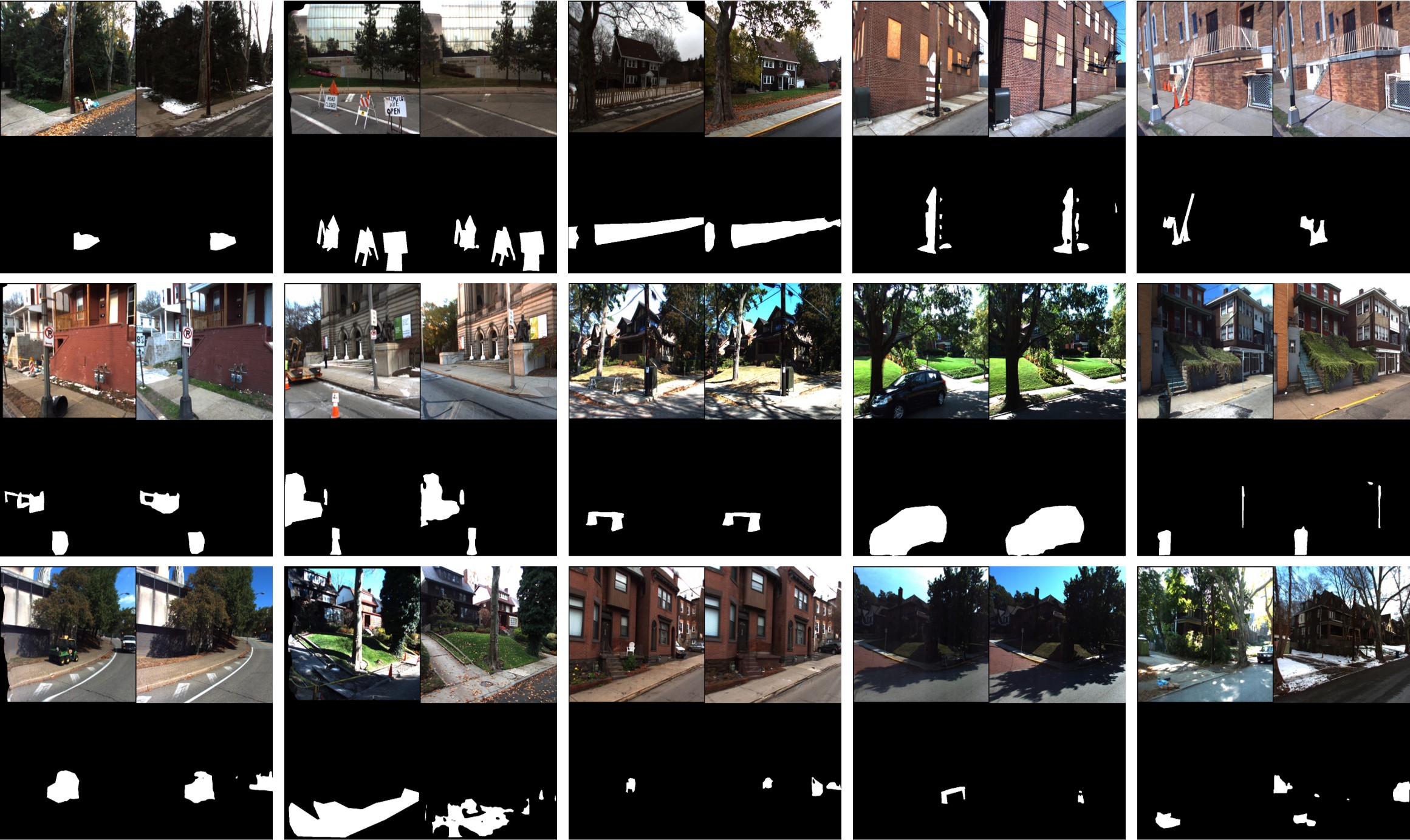}
	\caption{Visualization of C-3PO on the testing set of VL-CMU-CD. The last row shows some failure cases.}
	\label{fig:CMU}
\end{figure}

\begin{figure}
	\centering
	\includegraphics[width=\linewidth]{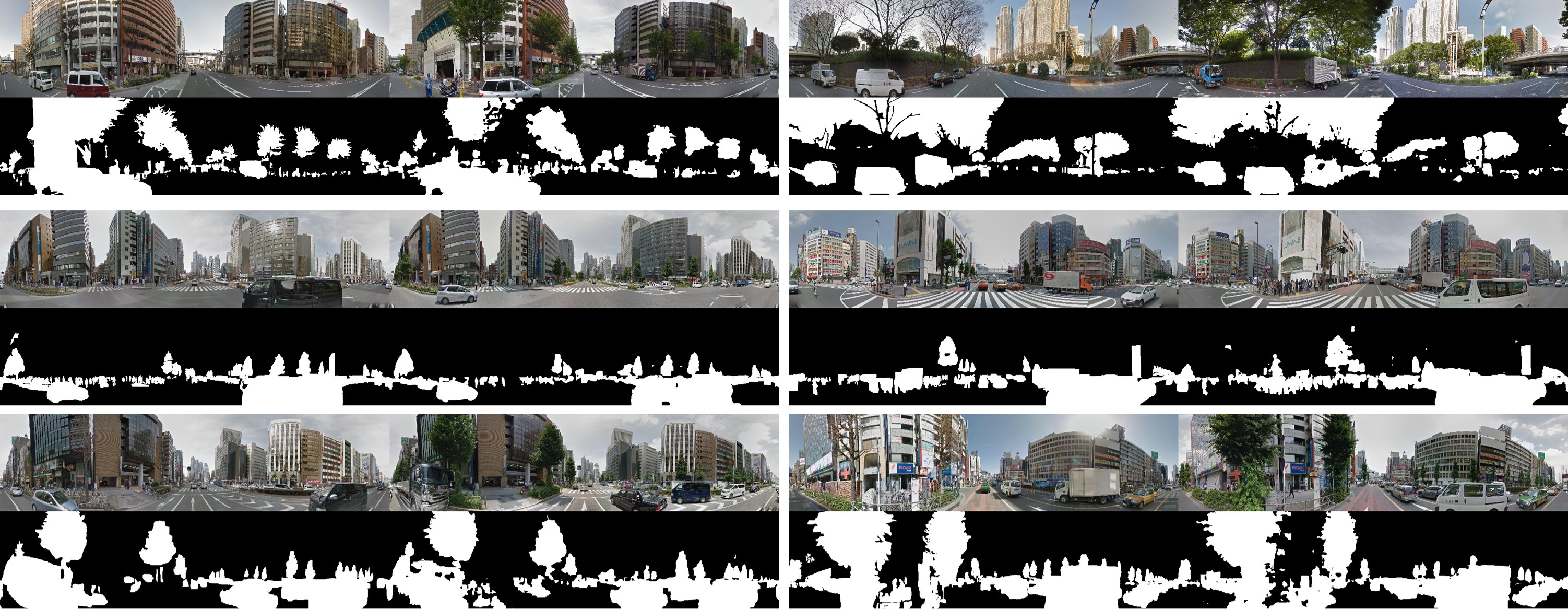}
	\caption{Visualization of C-3PO on the testing set of GSV.}
	\label{fig:GSV}
\end{figure}

\begin{figure}
	\centering
	\includegraphics[width=\linewidth]{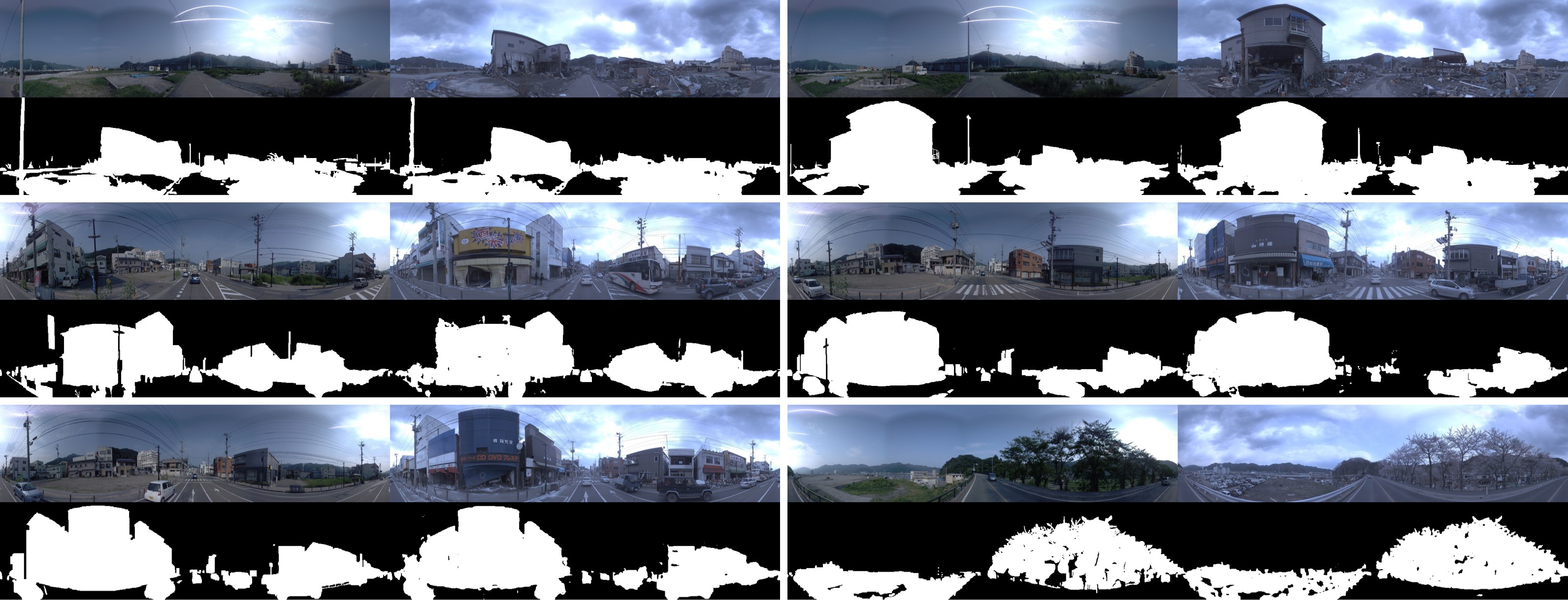}
	\caption{Visualization of C-3PO on the testing set of TSUNAMI.}
	\label{fig:TSUN}
\end{figure}

\begin{figure}
	\centering
	\includegraphics[width=\linewidth]{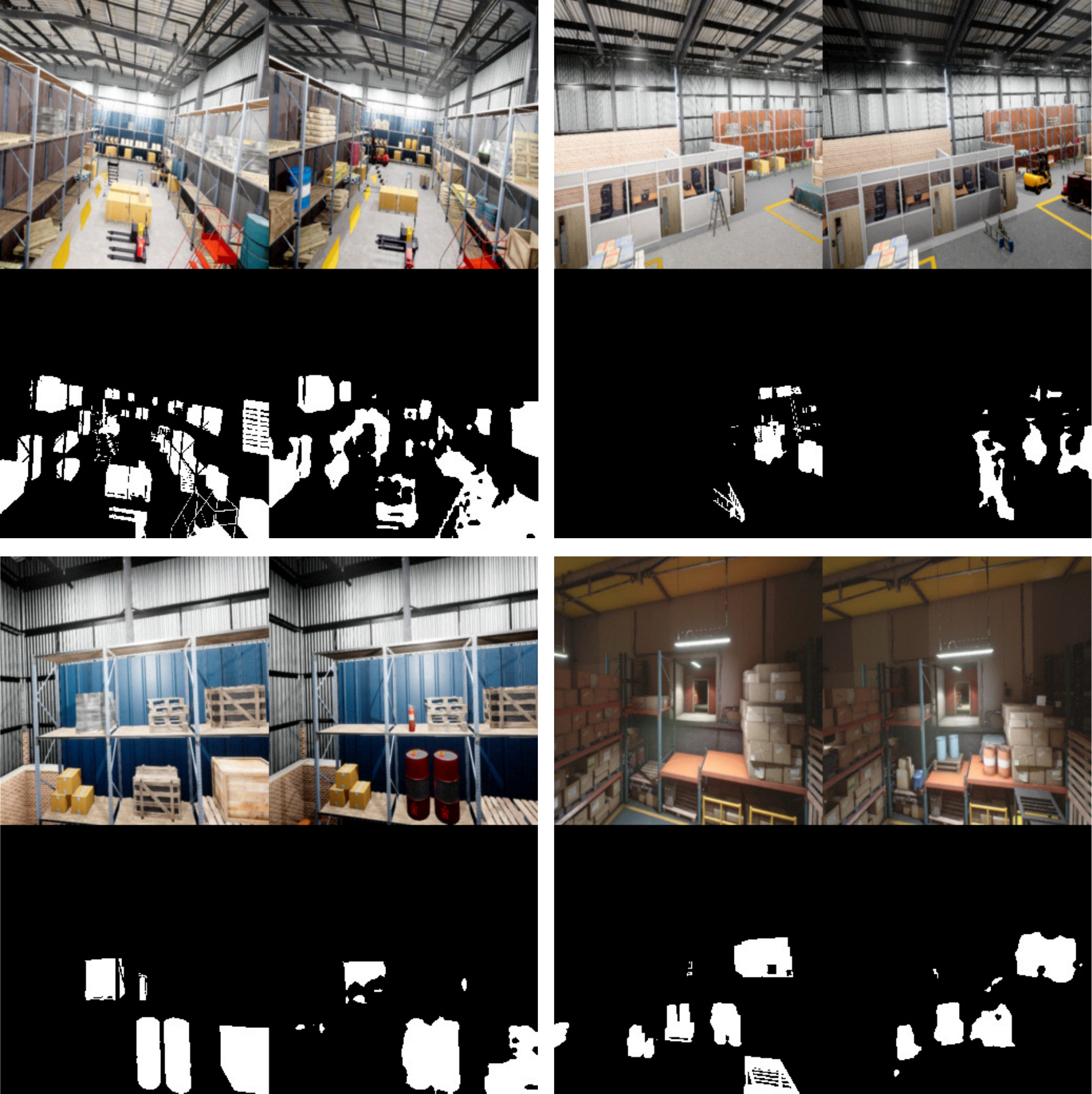}
	\caption{Visualization of C-3PO on the testing set of ChangeSim.}
	\label{fig:changesim}
\end{figure}

We visualize the predicted masks of C-3PO. Figs.~\ref{fig:CMU}, \ref{fig:GSV}, \ref{fig:TSUN}, and \ref{fig:changesim} illustrate the results of VL-CMU-CD, GSV, TSUNAMI, and ChangeSim, respectively. In each sub-figure, the top two images are the images captured at $t_0$ and $t_1$, respectively. The bottom two masks are the ground truth and the model's prediction, respectively.

\section{Conclusions and Future Work}

To solve the change detection problem, we proposed a new paradigm that reduces CD to semantic segmentation. Our framework decouples the CD parts and the segmentation parts. Directly applying the mainstream semantic segmentation networks help us relieve from the general segmentation problems in the CD task. And we only need to study how to fuse features for change detection. We proposed the MTF module to achieve this target. MTF is designed based on our insight of CD, that is the fusion features needs to contain the information of three possible change types. Finally, we proposed a C-3PO network for change detection. C-3PO is simple but effective. And it achieves state-of-the-art performance without bells and whistles.

With our paradigm, applying a more powerful semantic segmentation network is a promising way to further boost the performance. And how to identify the specific change object is an important problem in real-world applications. We leave these as future works.

\bibliography{egbib}

\end{document}